\documentclass[letterpaper, 10 pt, conference]{ieeeconf}  

\IEEEoverridecommandlockouts                              

\newcommand{\bx}{\mathbf{x}}

\usepackage[utf8]{inputenc}
\usepackage[T1]{fontenc}
\usepackage[sc]{mathpazo}
\usepackage[american]{babel}
\usepackage[autostyle]{csquotes}
\usepackage[%
  backend=bibtex,
  url=false,
  doi=false,
  style=numeric,
  giveninits,
  sorting=none,
  maxnames=10
]{biblatex}
\usepackage{booktabs} 
\usepackage[final]{microtype} 
\usepackage[hidelinks, breaklinks = true]{hyperref} 

\usepackage{amsmath}
\usepackage{algorithm} 
\usepackage[noend]{algpseudocode} 
\usepackage{units} 

\usepackage{graphicx}
\usepackage{psfrag}
\usepackage{tikz}

\usepackage{comment}	
\usepackage{acronym}
\usepackage{mathrsfs}
\usepackage{subcaption}
\usepackage{amsfonts}
\usepackage[capitalise]{cleveref}
\usepackage{hyperref}

\usepackage{bbm}

\bibliography{literature}

\title{\LARGE \bf
Density Planner: Minimizing Collision Risk in Motion Planning with Dynamic Obstacles using Density-based Reachability$^*$} 

\author{Laura Lützow$^{1,2}$, Yue Meng$^{2}$, Andres Chavez Armijos$^{3}$  and Chuchu Fan$^{2}$ 
\thanks{*Laura Lützow was supported by a fellowship within the IFI program of the German Academic Exchange
Service (DAAD). Ford Motor Company provided funds to assist the authors with their research, but this article solely reflects its authors' opinions and conclusions and not Ford's.}
\thanks{$^{1}$ Department of Informatics, Technical University of Munich, Germany}%
\thanks{$^{2}$ Department of Aeronautics and Astronautics, Massachusetts Institute of Technology, USA}%
\thanks{$^{3}$ Division of Systems Engineering, Boston University, USA}%
}

\begin{document}

\maketitle

\begin{abstract}
Uncertainty is prevalent in robotics. Due to measurement noise and complex dynamics, we cannot estimate the exact system and environment state. Since conservative motion planners are not guaranteed to find a safe control strategy in a crowded, uncertain environment, we propose a density-based method. Our approach uses a neural network and the Liouville equation to learn the density evolution for a system with an uncertain initial state. We can plan for feasible and probably safe trajectories by applying a gradient-based optimization procedure to minimize the collision risk. We conduct motion planning experiments on simulated environments and environments generated from real-world data and outperform baseline methods such as model predictive control and nonlinear programming. While our method requires offline planning, the online run time is 100 times smaller compared to model predictive control.

The code and supplementary material can be found at \url{https://mit-realm.github.io/density_planner/}. 
\end{abstract}


\section{INTRODUCTION}
\label{sec:introduction}

Ensuring safety is crucial for autonomous systems. 
However, because of uncertainties such as measurement errors, external disturbances, and model errors, we cannot precisely predict the future state of the system or the environment, which complicates safety certification. 

To deal with these uncertainties, there are two main directions in motion planning: Conservative approaches focus on robust safety by considering all possible situations via reachability analysis~\cite{althoff2010reachability}, planning a worst-case safe trajectory~\cite{lofberg2003approximations} or computing safe sets in which safety can be guaranteed for contained trajectories~\cite{rakovic2005invariant, gruber2020computing}. However, in many situations, a guaranteed safe trajectory may not exist. 
Thus, other approaches seek trajectories that are safe with high probability~\cite{mesbah2016stochastic,wang2020non} by assuming linear dynamics or Gaussian distributions to propagate the uncertainties along the trajectories. Our method follows a similar philosophy as the latter approach but we handle nonlinear dynamics and the propagation of arbitrary initial state uncertainties by leveraging machine learning techniques.

Recent work~\cite{article:DensityReach} solves the Liouville equation~\cite{article:LE} with a neural network to learn the density distribution of reachable states.
The follow-up work~\cite{meng2022case} conducts perturbation-based motion planning to ensure probabilistic safety. However, the method in~\cite{meng2022case} is computationally expensive and was not tested on realistic datasets. Inspired by this work, we design a differentiable framework to plan for feasible trajectories which minimize the collision risk. Beyond the system settings in~\cite{meng2022case}, we also model the uncertainty from the environment and estimate the collision probability in an efficient manner. The proposed planning process can be described as follows: First, we predict the evolution of the state distribution in time for a closed-loop system under a parameterized control policy using a neural network. Then, we model the environment uncertainty as a probability occupancy grid map. Finally, we use a gradient method to optimize the policy parameters by minimizing the collision risk.

We conduct experiments on both simulated and real-world data~\cite{bock2020ind} where our method can outperform a standard and a tube-based model predictive control (MPC) with a lower failure rate and lower online runtime. 
Furthermore, our method can be used for nonlinear system dynamics and uncertain initial states following an arbitrary probability distribution and can find plausible trajectories even when the worst-case collision is inevitable.

Our contributions are threefold: 
(1)~To the best of our knowledge, we are the first to propose a density-based, differentiable motion planning procedure for nonlinear systems with state and environment uncertainties. 
(2)~We provide an efficient method to compute the collision probability for closed-loop dynamics and an effective gradient-based algorithm for probabilistic safe trajectory planning. 
(3)~We test our approach on simulated and real-world testing cases where we outperform state-of-the-art motion planning methods.

\section{RELATED WORKS}
\label{sec:related-works}
{\textbf{Motion planning under uncertainties:}} Motion planning aims to find a trajectory that minimizes some cost function,  fulfills the kinematic constraints, and avoids collisions~\cite{latombe2012robot}.
Environment uncertainties can be treated as occupancy maps \cite{lee2019collision,francis2017stochastic,macek2003approach} where probability navigation functions~\cite{hacohen2019probability}, chance-constrained RRT~\cite{aoude2013probabilistically}, or constrained optimization techniques~\cite{huang2020safe} are applied to ensure safety.
However, most of these methods assume that the initial state is precisely known, but this is often not given in the real world due to sensor noise. Thus, it is desirable to consider the whole initial state distribution for motion planning as it is done by density-based planning methods.

{\textbf{Density-based planning:}} Many density-based planning approaches assume that the initial state follows a Gaussian distribution. Under linear dynamics, the mean and covariance of the state distribution can be easily propagated through time~\cite{chen2015optimal}, and the related optimal control problems can be solved by convex programming~\cite{okamoto2019optimal}. There exist several publications dealing with the covariance steering problem for nonlinear systems~\cite{ridderhof2019nonlinear,chen2021covariance,yi2020nonlinear}, but these methods are not directly applicable to non-Gaussian cases.

Controlling arbitrary initial distributions is closely related to optimal transport theory~\cite{krishnan2018distributed}. The work~\cite{caluya2020finite,caluya2021reflected} finds the optimal policy by solving the Hamilton-Jacobi-Bellman (HJB) equation. However, this method can only be used for full-state feedback linearizable systems and the convergence cannot be guaranteed. In~\cite{chen2019duality}, a primal-dual optimization is proposed to iteratively estimate the density and update the policy with the HJB equation, but this approach requires a time-consuming state discretization and cannot generalize to high-dimensional systems.

The closest work to ours is~\cite{meng2022case}, which estimates the density for nonlinear systems and arbitrary initial distributions with neural networks (NN). With a perturbation method and nonlinear programming, a valid control policy is searched. However, the planning algorithm is very near-sighted, the collision-checking procedure is still computationally expensive, and no optimality statements can be claimed. 
Our approach overcomes these drawbacks by utilizing a more efficient collision computation method and a gradient-based optimization algorithm to minimize the collision probability.

\section{PRELIMINARIES}
\label{sec:preliminaries}

{\textbf{Density prediction:}}
First, we consider an autonomous system $\dot{\bx}=f(\bx)$ where $\bx\in\mathbb{R}^n$ is the system state with initial density distribution $\rho_0(\bx)$ and $\rho(\bx, t)$ is the density of $\bx$ at time $t$. According to~\cite{article:LE}, $\rho(\bx, t)$ follows the Liouville equation (LE): 
\begin{equation}
\frac{\partial \rho}{\partial t}+\nabla \cdot (f\cdot \rho)=0,
\end{equation}
where $\nabla \cdot (f\cdot \rho)=\sum\limits_{i}^n \left(\frac{\partial}{\partial \bx_i}f_i(\bx)\rho(\bx,t)\right)$.  Following~\cite{chen2019duality}, we know that the density at the future state $\Phi(\bx_0, t)$ is
\begin{equation}
    \rho(\Phi(\bx_0, t), t) = \rho_0(\bx_0) \exp \bigl(\underbrace{ -\int_0^t \nabla \cdot f\left(\Phi\small(\bx_0, \tau\small)\right) d\tau }_{g(\bx_0, t)}\bigr),
    \label{eq:liou}
\end{equation}
where $\Phi$ denotes the flow map of the system.
By solving Eq.~\eqref{eq:liou}, which can be accelerated by approximating $g(\bx_0, t)$ with a neural network~\cite{meng2022learning}, and using interpolation methods, 
the whole density distribution $\rho(\cdot, t)$ at time $t$ can be estimated.

\textbf{{Problem formulation:}}
In this work, we want to define an optimal control policy that steers an autonomous system to a goal state while minimizing the collision probability with dynamic obstacles. The state and input constraints must be satisfied, and the system dynamics can be nonlinear in the states. 

In contrast to standard motion planning problems, we consider an uncertain initial state following an arbitrary but known probability density function. 
During the execution of the control policy, the state can be measured and used for closed-loop control.

\section{DENSITY-BASED MOTION PLANNING}
\label{sec:probability-computation}

In this section, we explain the proposed algorithm. First, we show how the density distribution of the closed-loop dynamics can be predicted given a parameterized control policy.
Based on this, we provide an algorithm to compute the collision probability.
Next, we introduce the cost function to evaluate the policy parameters. Lastly, we describe the gradient-based optimization method.

\subsection{Density Estimation for Controlled Systems} 
\label{sec:density-estimation-with-nn}

A controlled system $\dot{\mathbf{x}}=f(\mathbf{x}, \mathbf{u})$ with policy $\mathbf{u}=\pi(\mathbf{x}; \mathbf{p})$ and parameters $\mathbf{p}$ can be treated as an autonomous system with dynamics $f_{\mathbf{p}}(\mathbf{x})$, and hence, its density can be computed with Eq.~\eqref{eq:liou}. 

Although our approach can adapt to arbitrary control policies, we choose $\mathbf{p}$ to parameterize a reference trajectory that is tracked by a tracking controller. By utilizing contraction theory \cite{article:contractionCriteria} for the controller synthesis, the tracking error between the system and the reference trajectory can be upper-bounded and formal convergence guarantees can be provided. In this paper, the contraction controller is learned with the algorithm from~\cite{article:learningCCM} which requires the system to be control-affine. However, the remaining parts of the density planner do not need this special system structure such that the planner can be applied to arbitrary system dynamics when replacing the controller.

Analog to~\cite{meng2022learning}, we train an NN to approximate $g$ and $\Phi$ from \cref{eq:liou} using the initial state $\mathbf{x}_0$, the prediction time $t_k$, and the control parameters $\mathbf{p}$ as inputs.

Assuming that the initial density $\rho_0$ at state $\mathbf{x}_0$ is known, the density ${\rho}(\Phi(\bx_0, t_k), t_k)$ can be predicted.

\subsection{Computation of the Collision Probability}
\label{sec:compute-collision-probability}

\begin{figure}[hbt]
    \centering
    \includegraphics[page=5, width=0.47\textwidth, height=0.96\textheight, keepaspectratio]{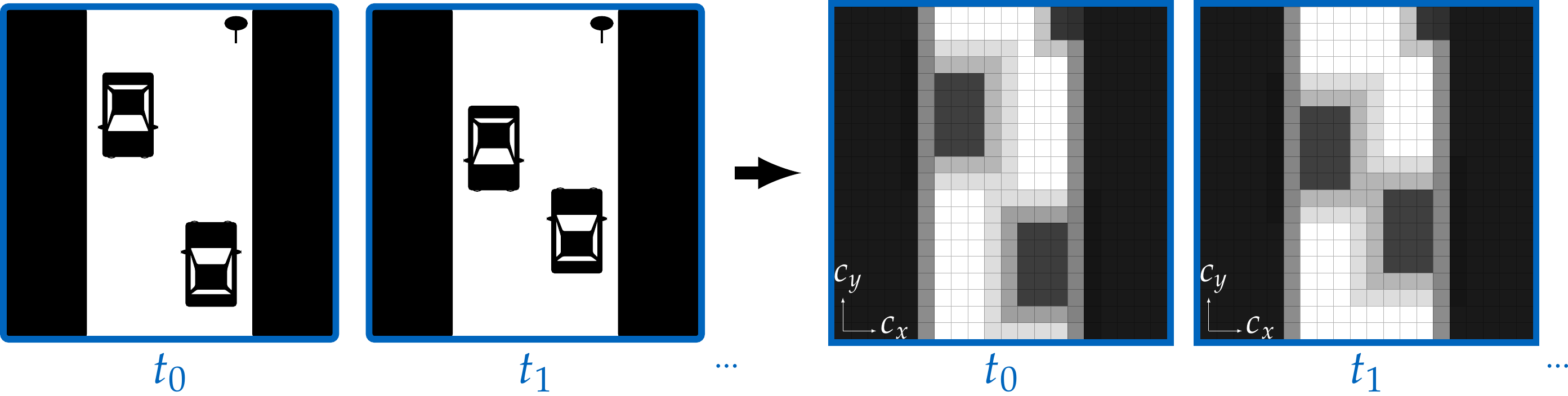}
    \caption{Discretization of the environment.} \label{fig:discrEnv}
    \vspace*{-\baselineskip}
\end{figure}

In this paper, we assume the probabilistic predictions for the evolution of a 2D environment are given as occupancy maps. The environment is evenly split into $C_x\times C_y$ grid cells along $x$ and $y$ directions. $P_{occ}(c_x, c_y, t_k)$ denotes the probability that cell $(c_x, c_y)$ is occupied by an obstacle at time $t_k$. Obstacles can be road users, lanes, or static objects, and their respective occupancy probabilities can be generated by off-the-shelf environment predictors as in~\cite{hoermann2018dynamic,wu2018probabilistic,koschi2018set}. The discretization of the environment is illustrated in \cref{fig:discrEnv} as a coarse grid map for visualization purposes.

We can compute the density distribution of the position of our ego vehicle as follows:  First, we randomly sample initial states $\{\bx^{(i)}(0)\}_{i=1}^N$ from the support of the given initial density distribution and predict the density $\{\hat{\rho}(\bx^{(i)}(t_k), t_k)\}_{i=1}^N$ along the reference trajectory $\mathbf{p}$ with the trained NN.
To compute the occupation probabilities of the ego vehicle
$P_{ego}(c_x, c_y, t_k)$ for every cell $(c_x, c_y)$ and timestep $t_k$, 
we assign to each predicted sample position $\hat{\bx}^{(i)}(t_k)$ a corresponding cell $(c^{(i)}_x, c^{(i)}_y)$ on the occupancy map. The densities of all samples falling into the same cell are averaged to represent the intermediate occupancy logits. Finally, the occupancy logits are normalized on the $xy$-plane to transform to the occupancy probability of the ego vehicle $P_{ego}$. The collision probability on $(c_x, c_y)$ can be computed as: 
\begin{align}
    P_{coll}(c_x, c_y, t_k) = P_{occ}(c_x, c_y, t_k) \cdot P_{ego}(c_x, c_y, t_k).
\end{align}
By integrating $P_{coll}(c_x, c_y, t_k)$ over the $x$ and $y$ dimensions, we can derive the overall collision probability at time $t_k$. For simplicity, we do not consider the physical dimensions of the system when predicting the future density distribution but assume a point shape.

\subsection{Cost Function Formulation}\label{sec:cost}
After computing the collision probability for a given reference trajectory $\mathbf{p}$, we seek to find the reference trajectory that reaches a target goal and stays within the valid state space while minimizing the collision probability and the control effort. On that account, we create a differentiable cost function 
as a convex combination of four sub-objectives with weighting factors $\alpha$ 
to evaluate trajectory $\bigl(\mathbf{x} (\cdot),~\mathbf{u} (\cdot)\bigr)$:
\begin{align} \label{eq:cost}
    J\bigl(\mathbf{x} (\cdot),\mathbf{u} (\cdot)\bigr) = &\alpha_{{G}} J_{{G}} + \alpha_{{I}} J_{{I}} 
    + \alpha_{{B}} J_{{B}} + \alpha_{{C}} J_{{C}}.
\end{align}
$J_G$ penalizes the distance from the final state of the state trajectory to the goal state, weighted by the final density of the trajectory,
\begin{equation}
    J_G = \rho(\mathbf{x}(t_N), t_N) ||\mathbf{x}(t_N) - \mathbf{x}_G||^2_{\mathbf{Q}_G}~,\label{eq:Jgoal}
\end{equation}
where $t_N$ is the final time, $\mathbf{x}_G$ is the goal state for the ego vehicle to reach and $\mathbf{Q}_G$ is a weighting matrix.
Similarly, $J_{I}$ penalizes the control effort,
\begin{equation}
J_{I} =  \sum_{k=0}^{N-1} ||\mathbf{u}(t_k)||^2_{\mathbf{Q}_{{I}}} ~ .
\label{eq:Jinput} 
\end{equation}
The term $J_{B}$ aims to keep the state trajectory within the valid state space and is defined as
\begin{align}
    J_{B} = \sum_{k=0}^{N} \rho(\mathbf{x}(t_k), t_k) \Bigl( &||\mathbf{x}(t_k) - \mathbf{x}_{{min}}||^2_{\mathbf{Q}_{{min}}(\mathbf{x}(t_k))}\notag\\
     + & ||\mathbf{x}(t_k) - \mathbf{x}_{{max}}||^2_{\mathbf{Q}_{{max}}(\mathbf{x}(t_k))}\Bigr)  \label{eq:Jbounds}
\end{align}
where $\mathbf{x}_{{min}}$ is the minimum and $\mathbf{x}_{{max}}$ is the maximum bound for the state space, and $\mathbf{Q}_{{min}}(\mathbf{x}(t_k))$ and $\mathbf{Q}_{{max}}(\mathbf{x}(t_k))$ are diagonal matrices whose diagonal elements are nonzero only when the corresponding state exceeds the maximum or falls below the minimum bound, respectively.

The last term, $J_{C}$, describes the cost for high collision probabilities. Due to the non-differentiability of the implemented binning method for computing $P_{ego}$, 
the collision probability cannot be minimized directly by a gradient-based optimization algorithm.
Instead we compute a differentiable collision risk for $\mathbf{x}(t_k)$ as follows:
\begin{enumerate}
    \item We approximate the gradients of $P_{occ}$ between adjoining cells in $x$ and $y$ direction denoted by $G_x$ and $G_y$, respectively.
    \item Following the negative gradient, we can  compute the desired position for sample $\mathbf{x}(t_k)$:
    \begin{align}
        c_{{x}, \text{des}}(t_k) &= c_{x}(t_k) - \beta G_{x}(c_x, c_y, t_k) \label{eq:cx_des}\\
        c_{y, \text{des}}(t_k) &= c_{y}(t_k) - \beta G_{y}(c_x, c_y, t_k),\label{eq:cy_des}
    \end{align}
    where $\beta$ is the step size, and $c_x$ and $c_y$ are the $x$ and $y$ position of $\mathbf{x}(t_k)$ in grid coordinates, respectively. 
    \item     After transforming the desired grid position to the position in real-world coordinates $(p_{{x}, \text{des}}, p_{{y}, \text{des}})$, 
    we can compute the collision risk cost as
    \begin{align}
        J_{C} = \sum_{k=0}^{N} P_{coll}&(\mathbf{x}(t_k)) \Bigl(\bigl(p_{x}(t_k)-p_{{x}, \text{des}}(t_k)\bigr)^2 \notag \\+ &\bigl(p_{y}(t_k)-p_{{y}, \text{des}}(t_k)\bigr)^2\Bigr), \label{eq:Jcoll}
    \end{align} 
    where the collision probability for the reference trajectory is computed as $P_{coll}(\mathbf{x}(t_k)) = P_{occ}(\mathbf{x}(t_k), t_k) \rho(\mathbf{x}(t_k), t_k)$.
    The minimization of $J_{C}$ yields positions $\mathbf{x}(t_k)$ with low occupancy probability, and consequently, decreases the collision probability of the trajectory.
\end{enumerate}
Finally, the overall cost function can be minimized by gradient descent methods such as ADAM~\cite{article:AdamAM}.


\subsection{Optimization Approach} \label{sec:optAppr}
We propose the following two-step procedure for optimizing the reference trajectory:


\textbf{{Initialization:}}
Since the optimization of \cref{eq:cost} for an uncertain initial state is computationally complex and due to its non-convexity can get stuck in local minima, we wish to find a suitable initial guess first without performing any density predictions.
This can be done by assuming an exactly given initial state which is chosen as the mean of the initial density distribution and optimizing a large number of reference trajectories starting from there. The process is given as follows: 

First, we randomly sample $M$ parameter sets $\{\mathbf{p}^{(i)}\}_{i=1}^M$ from the admissible policy parameter space and recover the corresponding reference trajectories starting at the predefined initial state. 
Next, we calculate the cost with \cref{eq:cost} for each parametrization by assuming that the system follows the corresponding reference trajectory accurately. We update the parameters $\{\mathbf{p}^{(i)}\}_{i=1}^M$ with gradient descent until a certain number of iterations is reached. Since we do not consider the uncertainty of the initial state, we assume that the density of each state is equal to one.

Given that we wish to guide the trajectories to the goal, we initialize $\alpha_{B}$ and $\alpha_{{C}}$ in \cref{eq:cost} for all trajectories with zero. 
At each optimization iteration and for each trajectory pair, we check if the distance to the goal is smaller than a  threshold. If this is true for one trajectory, the corresponding $\alpha_{B}$ is set to a nonzero value such that the state space constraints get enforced.
For all trajectories where $\alpha_{B} \not= 0$, we perform another check: If the state space constraints are met, the collision cost will be considered by setting $\alpha_{{C}}$ to a nonzero value.
This cost calculation procedure ensures that the trajectories are first steered to the goal, are next moved to the valid state space, and only then get optimized with respect to collisions. This procedure saves computation effort and leads to good trajectories much faster than when considering all cost terms from the beginning.
Finally, we compare the costs of all optimized parameter sets and return the set $\mathbf{p}^*$ with the lowest cost.


\textbf{{Local optimization with density predictions:}} 
In the next step, we locally optimize the best parameter set $\mathbf{p}^*$ by taking the initial state uncertainties into account. 
First, we randomly sample $S$ initial states $\{\mathbf{x}^{(i)}(0)\}_{i=1}^S$ from the given initial density distribution $\rho(\cdot, 0)$ and approximate their state trajectories $\mathbf{x}^{(i)}(\cdot)$ as well as their density $\rho(\mathbf{x}^{(i)}(\cdot), \cdot)$ with the neural density predictor when tracking reference trajectory $\mathbf{p}^*$.

Next, we compute the costs $J^{(i)}$ of all trajectories with \cref{eq:cost} where all weights $\alpha$ are nonzero from the beginning.
If the maximum number of optimization iterations is not reached, we calculate the gradient of the overall cost $\sum_{i=1}^S J^{(i)}$ and optimize $\mathbf{p}^*$.

\section{EXPERIMENTS}\label{sec:applications}
While the proposed motion planning algorithm can be applied to all kinds of autonomous systems, we implement our method on a self-driving car in a congested uncertain environment. In this section,
we introduce the model and problem for an autonomous-driving scenario. Additionally, we perform an ablation study on the proposed optimization method. 
Lastly, we compare our algorithm against three baseline methods in different environments. 
All experiments were conducted on an AMD Ryzen Threadripper 3990X (2.9GHz) with an NVIDIA A4000 16GB graphics card. The environments considered in the experiments are visualized in the supplementary video.

{\bf Model description: }
The system is described by a kinematic vehicle model \cite{article:dubins} with input $\mathbf{u}(t)=[\omega(t), a(t)]^T$, where $\omega(t)$ is the angular velocity and $a(t)$ is the longitudinal acceleration. Additionally, to demonstrate the robustness of the proposed algorithm to disturbances, we included a constant sensor bias ${\theta}_\text{bias}$ for measuring the heading angle. The resulting vehicle dynamic model is defined as
\begin{equation}
    \dot{\mathbf{x}}(t) = \begin{bmatrix} \dot{p}_{x}(t)\\ \dot{p}_{y}(t)\\ \dot{\theta}(t)\\ \dot{v}(t) \\ \dot{\theta}_\text{bias}(t) \end{bmatrix}
        = \begin{bmatrix} v(t) ~\cos{(\theta(t)) }\\ v(t) ~\sin{(\theta(t))}\\ \omega(t)\\ a(t) \\ 0\end{bmatrix}
\end{equation} 
where $p_{x}(t)$ and $p_{y}(t)$ describe the position along the $x$-axis and $y$-axis respectively. $\theta(t)$ denotes the heading angle and $v(t)$ the longitudinal velocity. 
The control input $\mathbf{u}$ is generated by the learned contraction controller to track the reference trajectory on the basis of a biased state measurement $\mathbf{\hat{x}}(t) = [p_{x}(t),~ p_{y}(t),~ \theta(t)+\theta_\text{bias}(t),~ v(t),~ 0]^T$.

\begin{figure}[pt]
    \begin{subfigure}[b]{.36\linewidth}
        \centering
        \includegraphics[width=\textwidth, height=0.29\textheight, keepaspectratio]{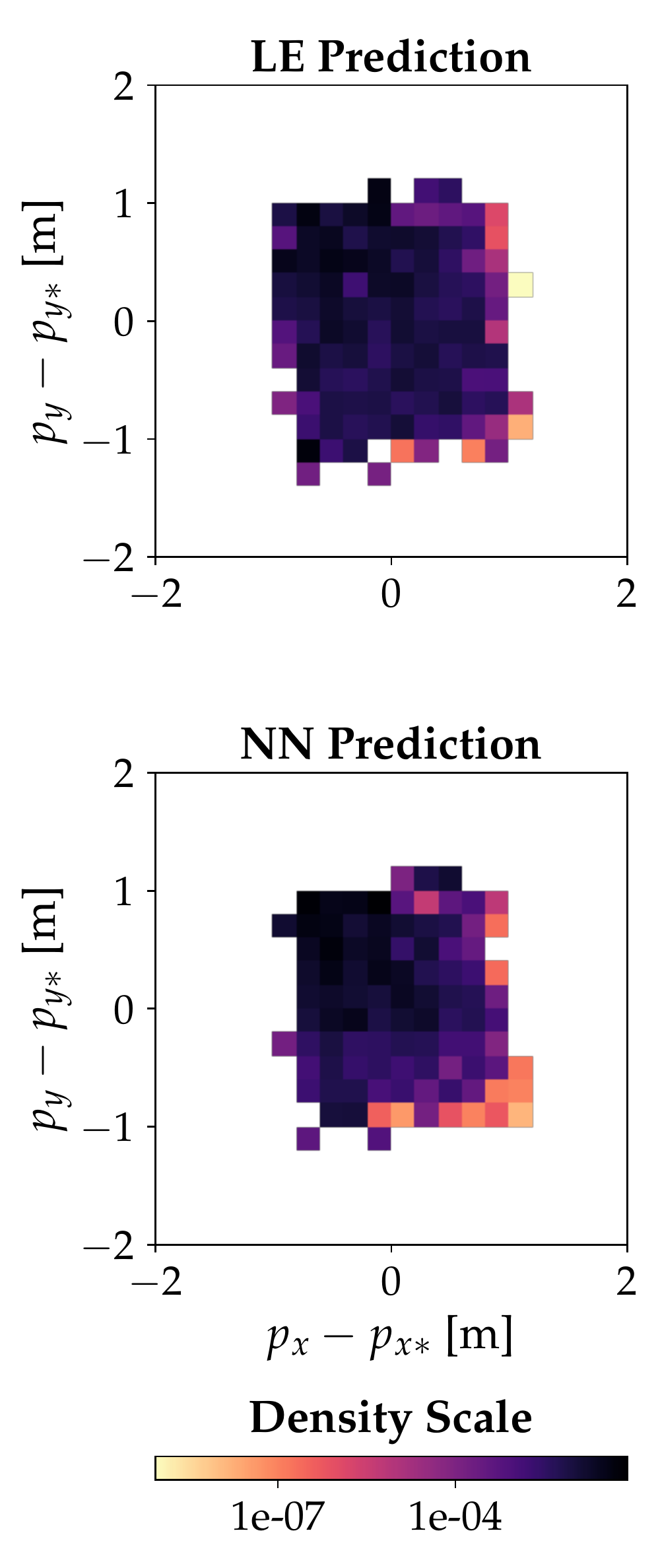}
        \vspace*{-2mm}
        \caption{$t_k=0.5s$}
        \label{fig:density_pred0}
    \end{subfigure}%
    \begin{subfigure}[b]{.3\linewidth}
        \includegraphics[width=\textwidth, height=0.29\textheight, keepaspectratio]{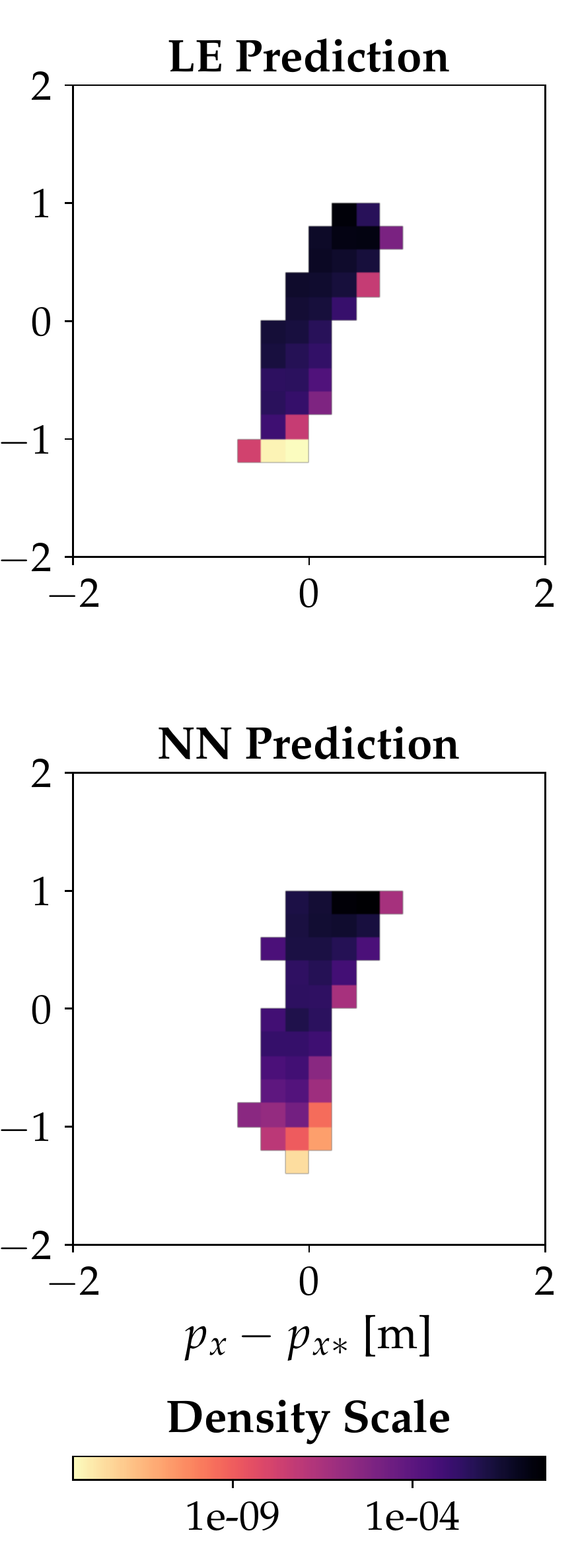}
        \vspace*{-6mm}
        \caption{$t_k=4s$}
        \label{fig:density_pred1}
    \end{subfigure}
    \begin{subfigure}[b]{.3\linewidth}
        \centering
        \includegraphics[width=\textwidth, height=0.29\textheight, keepaspectratio]{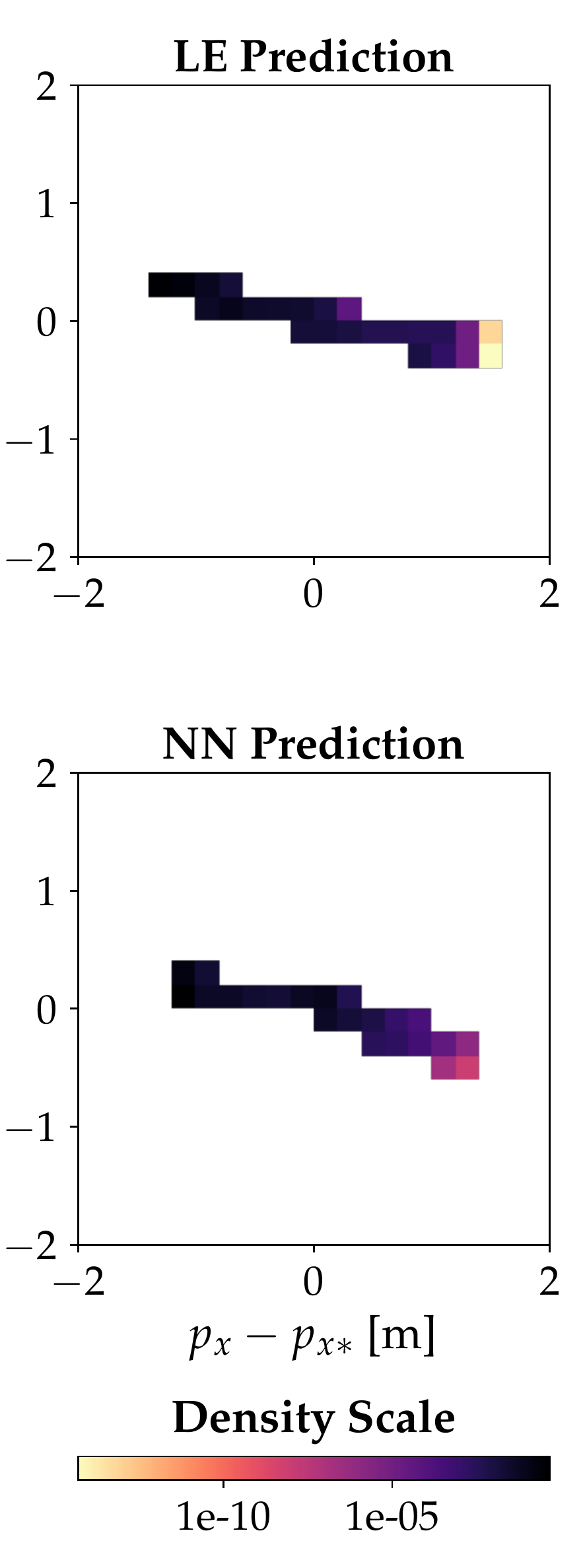}
        \vspace*{-6mm}
        \caption{$t_k=10s$}
        \label{fig:density_pred2}
    \end{subfigure}
    \caption{Performance of the density NN.}
    \vspace*{-\baselineskip}
    \vspace*{-3mm}
    \label{fig:PerformNN}
\end{figure}

Next, we train the NN to predict the density along the reference trajectory.
 For this purpose, we define an architecture with seven fully connected layers and 150 neurons. 
 A sample of the density predictions is shown in \cref{fig:PerformNN} where the NN predictions are compared against the solution of the LE as defined by \cref{eq:liou}.  
While the prediction of 10s-trajectories of the state and the density starting from 500 initial states takes on average 11.6s when integrating the system dynamics and using LE, the NN decreases this time to 0.3s. 

Lastly, to plan the reference trajectory, we define the following weighting matrices:
\begin{align*} 
    \mathbf{Q}_G = \text{diag}([1, 1,0,0,0]),~
    \mathbf{Q}_I = \text{diag}([1, 1]),
\end{align*}
where diag() denotes a diagonal matrix.
The cost factors $\alpha$ in \cref{eq:cost} are chosen such that the resulting cost gradient is sensible to $J_G$ and $J_C$ but is dominated by $J_B$ when the trajectory is out-of-bounds. By tuning $\alpha_G$ and $\alpha_C$, we can adjust the trade-off between low collision probabilities and goal-directed behavior. Furthermore, we clip the gradient if its absolute value becomes too big to ensure numerical stability.

\begin{figure}[hpt]
    \vspace*{-2mm}
    \centering
    \includegraphics[page=12, width=\linewidth, keepaspectratio]{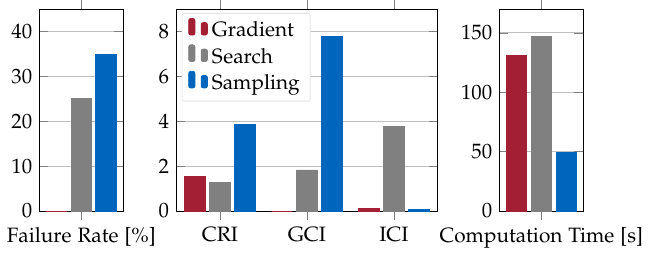}
    \vspace*{-5.5mm}
    \caption{Comparison of the optimization methods.} \label{fig:compOpt}
    \vspace*{-2mm}
\end{figure}

{\bf Trajectory optimization: }
To analyze the performance of the reference trajectory computation, we compare the proposed gradient-based optimization method with a sampling-based and a search-based approach. For this purpose, we generate environments with a random number of uncertain stationary and dynamic obstacles. Similarly, the obstacle sizes, positions, uncertainties, and velocities are chosen randomly.  
Furthermore, we define the following criteria for the evaluation of the aforementioned methods:
\begin{itemize}
    \item The \emph{failure rate} is the percentage of environments where the considered method did not find a solution within five minutes. Failure cases are not considered when computing the average computation time or the cost scores.
    \item The collision risk increase (\emph{CRI}), goal cost increase (\emph{GCI}), and input cost increase (\emph{ICI}) quantify the average increment of the collision risk, the goal cost, and the input cost when compared to the best possible method in each environment.
    \item The \emph{computation time} is averaged over all evaluations.
\end{itemize}
The results are illustrated in \cref{fig:compOpt}. It can be seen that the proposed gradient-based approach achieves better performance in almost every criterion as compared to the other two methods. In each scenario, the gradient-based method was able to provide a valid solution while the sampling-based method failed in 35\% of the environments. In addition, the distance from the final state of the planned trajectory to the goal was on average 0.2m as compared to the search-based (1.4m) and the sampling-based method (2.8m). 
Additionally, the collision risk of the gradient-based approach is comparatively small in most environments with an acceptable computation time.

{\bf Baseline methods: }
To evaluate the performance of the proposed density planner, we compare it against other collision-minimizing motion planning approaches. Specifically, we define two receding horizon MPC algorithms for the given motion planning task. 

The first implementation corresponds to an \emph{standard MPC}, denoted as \emph{M0}, which does not consider any disturbances or model uncertainties. The corresponding cost function is defined as:
\begin{align}
    J_\text{MPC} =& 
    \alpha_I \sum_{k=h}^{h + H-1} ||\mathbf{u}(t_k)||^2_{\mathbf{Q}_{{I}}}  +
    \alpha_G ||\mathbf{x}(t_{h + H}) - \mathbf{x}_G||_{\mathbf{Q}_G}   \notag\\ &+ \alpha_C \sum_{k=h}^{h + H} \bigl(P_{coll}(\mathbf{x}(t_k), t_k)\bigr)^2, \label{eq:optNLP} 
\end{align} 
where $H$ denotes the prediction horizon, $t_{h}$ is the current point in time, and $\alpha$ and $\mathbf{Q}$ are the same weighting factors and matrices as in \cref{eq:cost}.

Similarly, to handle imperfect measurements due to sensor bias, we define a \emph{tube-based MPC} which minimizes the collision probability and enforces the state space constraints within a tube around the nominal trajectory. 
We will test implementations with the tube radii $0.3m$, $0.5m$, and $1m$ which are denoted as \emph{M1}, \emph{M2}, and \emph{M3}, respectively.

For each environment, we assume that the minimum cost is achieved by an oracle, \emph{O}, that computes the optimal solution of Prob. \eqref{eq:optNLP} with $h=0$ and $H=N=100$. 
To compute the optimal solution, we do not consider computation time constraints nor uncertainties. Thus, we assume that the oracle knows the true initial state and the exact system model.

\textbf{Numerical evaluation: }
First, we analyze the motion planning methods in 50 randomly generated environments. The motion planners are compared using the same criteria defined in the ablation study with the difference that a failure is reported when $J_C>10$ or  when the distance from the final state to the goal is larger than 4.5m (the distance between the start and the goal position lies between 10 and 70m). 
\begin{figure}[hpt]
    \vspace*{-2mm}
    \centering
    \begin{subfigure}{0.23\textwidth}
        \centering
        \includegraphics[page=13, width=\textwidth, height=0.22\textheight, keepaspectratio]{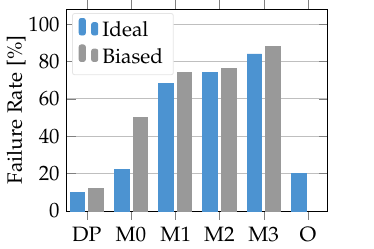}
        \vspace*{-5.5mm}
        \caption{Failure rate.} \label{fig:compMP_failure}
    \end{subfigure}
    \begin{subfigure}{0.23\textwidth}
        \centering
        \includegraphics[page=14, width=\textwidth, height=0.22\textheight, keepaspectratio]{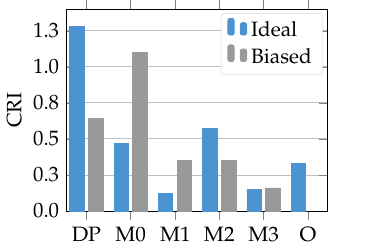}
        \vspace*{-5.5mm}
        \caption{Collision risk.} \label{fig:compMP_coll}
    \end{subfigure}
    \begin{subfigure}{0.23\textwidth}
        \centering
        \includegraphics[page=15, width=\textwidth, height=0.22\textheight, keepaspectratio]{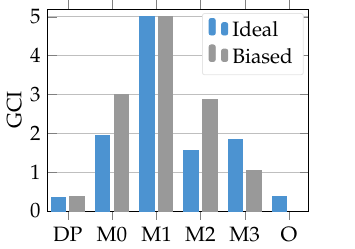}
        \vspace*{-5.5mm}
        \caption{Goal cost.} \label{fig:compMP_goal}
    \end{subfigure}
    \begin{subfigure}{0.23\textwidth}
        \centering
        \includegraphics[page=16, width=\textwidth, height=0.22\textheight, keepaspectratio]{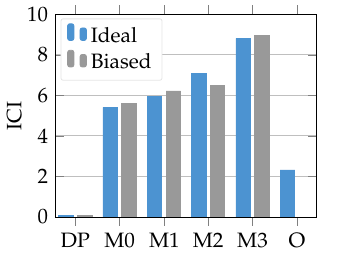}
        \vspace*{-5.5mm}
        \caption{Input cost.} \label{fig:compMP_input}
    \end{subfigure}
    \caption{Comparison in artificial environments.} \label{fig:compMP}
    \vspace*{-3mm}
\end{figure}

In \cref{fig:compMP_failure}, we can see that the proposed density planner (\emph{DP}) is the most reliable method given a success rate of 90\% when using perfect measurements. Furthermore, all trajectories end within an average distance of $0.71m$ from the goal and low input cost.
However, the collision risk in \cref{fig:compMP_coll} is higher when compared to the other motion planning approaches due to the higher risk tolerance of the proposed algorithm: 
Even in very crowded environments, DP is able to reach the goal and stay below the acceptance tolerance for the collision risk ($J_C \leq 10$) such that these environments are considered in the computations of the CRI. 
The other planners, on the other side, either reach the goal with a very low collision risk or are very far off such that a failure is reported and the trajectory is discarded.

When the measurements are biased, the failure rate of DP is marginally higher than for the perfect measurement cases. This shows that the proposed algorithm is robust against deterministic disturbances. The collision risk is, on average, lower compared to the case of perfect measurements since the density NN was mostly trained with data where $\theta_\text{bias} \not= 0$ and hence better approximates the density for these cases. In contrast, the MPC methods present lower collision risks at the expense of higher goal costs and higher failure rates due to the conservative nature of the algorithms. 
Additionally, the input costs are considerably higher than for {DP} which is a byproduct of the short prediction horizon. 

\begin{figure}[bt]
    \vspace*{2mm}
    \centering
    \begin{subfigure}{0.48\linewidth}
    \centering
    \includegraphics[page=17, width=1\textwidth, height=0.2\textheight, keepaspectratio]{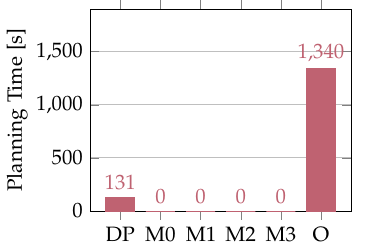}
    \vspace*{-5.5mm}
    \caption{Offline planning time.} \label{fig:compMP_offlineT1}
    \end{subfigure}
    \begin{subfigure}{0.48\linewidth}
    \centering
    \includegraphics[page=18, width=1\textwidth, height=0.2\textheight, keepaspectratio]{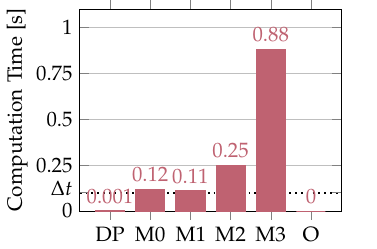}
    \vspace*{-5.5mm}
    \caption{Online computation time.} \label{fig:compMP_onlineT}
    \end{subfigure}
    \caption{Computational time evaluation.} \label{fig:compMP_T}
    \vspace*{-\baselineskip}
    \vspace*{-2mm}
\end{figure}

\cref{fig:compMP_offlineT1} shows the planning time which the algorithms need before starting the online control. DP takes on average $131s$ to compute the reference trajectory, which is a tenth of the planning time of the oracle. 
Furthermore, the oracle requires the true initial state when starting the planning while the density planner only needs the initial density distribution. The MPCs solve the corresponding optimization problems online and consequently do not need any planning time in advance.
On the other hand, the online computational complexity of the MPC algorithms is high, as shown in \cref{fig:compMP_onlineT}. Only M0 and M1 are able to solve the optimization problem in approximately real-time when using a time step of $\Delta t = 0.1s$. 
DP only has to compute the output of the contraction controller online to follow the precomputed reference trajectory. Hence, the resulting online computational complexity is significantly lower.

\textbf{Evaluation on a real-world dataset:}
Next, we show that the proposed algorithm can be applied to real-world environments without modifications. However, we assume access to the predicted occupancy maps for the environment. Here, we use the inD dataset \cite{bock2020ind} which contains a collection of naturalistic vehicle, bicyclist, and pedestrian trajectories recorded at German intersections by drones.
We add stochasticity around each traffic participant by using a Gaussian distribution with a mean equal to the vehicle position and a standard deviation equal to the obstacle length plus 1m. Additionally, we skewed the distribution along the direction of movement.

\begin{figure}[hpt]
    \vspace*{-3mm}
    \centering
    \includegraphics[page=19, width=0.35\textwidth, height=0.22\textheight, keepaspectratio]{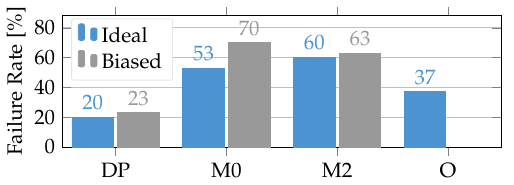}
    \vspace*{-2mm}
    \caption{Failure rate in real-world environments.} \label{fig:compMPR_fail}
    \vspace*{-\baselineskip}
\end{figure}

To evaluate the performance, we compare our algorithm against M0, M2, and O. 
We choose M2 since its tube radius is a compromise between robustness and low computational complexity. 
We randomly sample start and goal positions for ten random time periods at three intersections and compute the costs for each planner.
The failure rate is visualized in \cref{fig:compMPR_fail}. DP achieves the best results; it finds a path to the goal in 24 of 30 environments when using perfect measurements. 

\section{CONCLUSIONS AND FUTURE WORK}
\label{sec:conclusion}

In this paper, we have proposed a density-based motion planning approach to ensure system safety under uncertainties. The approach is based on a differentiable cost function that considers the density evolution starting from arbitrary initial state distributions and can handle nonlinear system dynamics.
The motion planner was applied to simulated environments generated from artificial data and from real-world data where it outperformed the baseline methods with significantly less online computational complexity.

However, the algorithm needs a planning time of approximately 130s to compute the reference trajectory on basis of the predicted environment occupations and the initial state distribution. As state-of-the-art environment predictors usually cannot look far ahead, the planning time must be decreased by using a more efficient implementation in C or C++ or by modifying our algorithm to perform in a receding horizon fashion. 
Furthermore, we plan to improve the density NN by using a more sophisticated architecture.
For instance, the usage of a multi-dimensional convolutional NN could enable the prediction of the whole density distribution at once such that a differentiable expression for the collision probability could be derived. Thus, instead of minimizing the collision risk for density-weighted sample trajectories, we could minimize the collision probability directly.
Additionally, we want to extend the approach to systems with stochastic disturbances by replacing the LE with the Fokker-Planck equation \cite{article:uncFPE, article:SolvingFPE}.




\section*{References}

\printbibliography[heading=none]{}






\end{document}